\documentclass{article} 
\usepackage{iclr2021_conference,times}

\usepackage[utf8]{inputenc} 
\usepackage[T1]{fontenc}    
\usepackage{hyperref}       
\usepackage{url}            
\usepackage{booktabs}       
\usepackage{amsfonts}       
\usepackage{nicefrac}       
\usepackage{microtype}      
\usepackage[disable]{todonotes}
\usepackage{wrapfig}
\usepackage{amsmath}
\usepackage{lipsum}  
\usepackage[linesnumbered,boxed]{algorithm2e}

\newcommand{\ca}[1]{\todo[inline]{\textbf{Cedric: }#1}}

\iclrfinalcopy

\newcommand{\R}{\mathbb{R}}
\newcommand{\HW}{\mathcal{H}}
\newcommand{\HP}{\mathcal{X}}
\newcommand{\ntask}{N}
\newcommand{\nobj}{m}
\newcommand{\nhp}{n}

\newcommand{\benchmark}[1]{\textit{#1}}
\newcommand{\fcnet}{\benchmark{fcnet}}
\newcommand{\fcnetcloud}{\benchmark{fcnet-cloud}}
\newcommand{\nas}{\benchmark{nas201}}
\newcommand{\nascloud}{\benchmark{nas201-cloud}}
\newcommand{\hwnas}{\benchmark{hw-nas}}

\newcommand{\HB}{HB}
\newcommand{\HBND}{+ND}
\newcommand{\HBNDtr}{+ND+tr}
\newcommand{\HBtr}{+tr}
\newcommand{\HBparego}{+Par}

\newcommand{\HBrw}{+RW}
\newcommand{\HBhv}{+HV}

\title{A multi-objective perspective on jointly tuning hardware and hyperparameters}

\author{%
  David Salinas
  \texttt{dsalina@amazon.com} \\
  Valerio Perrone
  \texttt{vperrone@amazon.com} \\
  Olivier Cruchant
  \texttt{cruchant@amazon.com} \\
  Cedric Archambeau
  \texttt{cedrica@amazon.com} \\  
}

\begin{document}

\maketitle

\begin{abstract}
In addition to the best model architecture and hyperparameters, a full AutoML solution requires selecting appropriate hardware automatically. 
This can be framed as a multi-objective optimization problem: there is not a single best
hardware configuration but a set of optimal ones 
achieving different trade-offs between cost and runtime. In practice, some 
choices may be overly costly or take days to train. To lift this burden, we adopt a multi-objective approach that selects and adapts the hardware configuration automatically alongside neural architectures and their hyperparameters.
Our method builds on Hyperband and extends it in two ways. First, 
we replace the stopping rule used in Hyperband by a non-dominated sorting rule to preemptively stop unpromising configurations. Second, we leverage hyperparameter evaluations from related tasks via transfer learning by building a probabilistic estimate of the Pareto front that finds promising configurations more efficiently than random search. We show in extensive NAS and HPO experiments that both ingredients bring significant speed-ups and cost savings, with little to no impact on accuracy. In three benchmarks where hardware is selected in addition to hyperparameters, we obtain runtime and cost reductions of at least 5.8x and 8.8x, respectively. 
Furthermore, when applying our multi-objective method to the tuning of hyperparameters only, we obtain a 10\% improvement in runtime while maintaining the same accuracy on two popular NAS benchmarks.
\end{abstract}

\section{Introduction}
\label{sec:introduction}
The goal of neural architecture search (NAS) is to automatically find well-performing deep neural networks 
in domains such as computer vision and natural language processing. While initial efforts focused on discovering configurations with high accuracy, further work extended NAS to optimize for inference latency which is key to enable embedded devices or smartphones \citep{Tan2018,Elsken2019}. Methodologies that tune configurations to optimize other criteria than accuracy for a fixed hardware have been referred to as hardware-aware NAS \cite[see][for a survey]{benmeziane2021comprehensive}. Recently, \cite{Wu2019} and \cite{li2021hwnasbench} proposed benchmarks that track latency in addition to accuracy for various hardware choices in order to compare hardware-aware approaches.

\begin{figure}
\center
\includegraphics[width=0.495\textwidth]{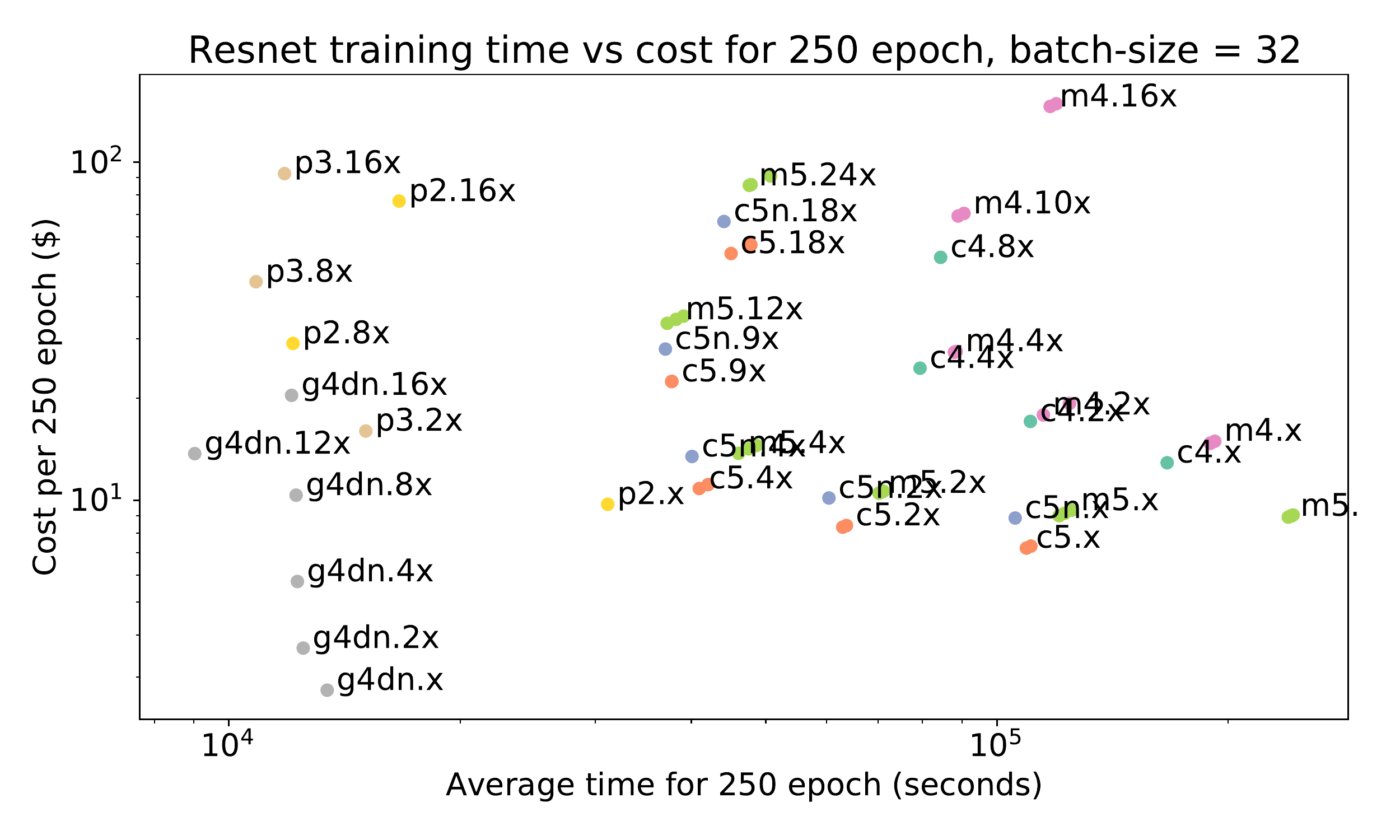}
\includegraphics[width=0.495\textwidth]{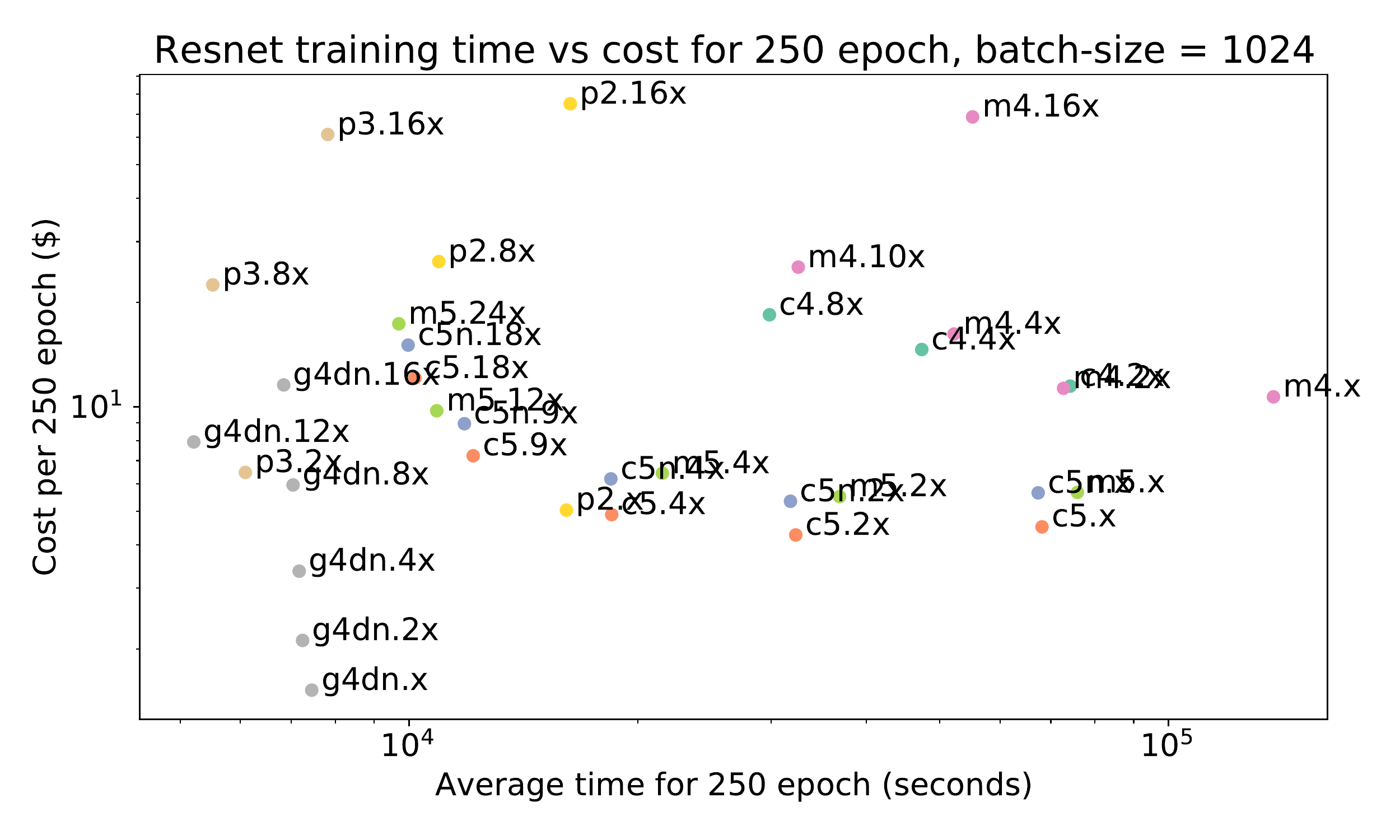}
\caption{Time and cost to train Resnet on Cifar10 on AWS instance-types for a small (left) and a larger batch-size (right). The axis are in a logarithmic scale and colors indicate instance families. It can be observed that time and cost span several orders of magnitude. \label{fig:hw-perf}}
\end{figure}

While much work focused on finding good architectures given a \emph{fixed} hardware target, to the best of our knowledge no work considered tuning hardware together with architectures (or hyperparameters) in this context. NAS often starts \emph{after} choosing a hardware configuration (e.g., the type and numbers of GPUs to use) which can have a massive impact on runtime and cost. Figure~\ref{fig:hw-perf} shows the measured runtime when training Resnet on Cifar10 with all the machines available on AWS for two different batch-sizes. This reveals two key aspects of the hardware selection problem. First, there is not a single best instance but a set of optimal trade-offs. For instance, for the large batch-size (right panel) the optimal instances are p3.2x and all g4dn instances as they are on the Pareto front, which means that using any other instance will be both more expensive and slower to train Resnet for 250 epochs and a batch size of 1024. Second, the performance is non-stationary with respect to the hyperparameters. Indeed, changing the batch-size causes the hardware performance to drastically change (the axis are logarithmic). An even stronger effect would be expected across different tasks, calling for methods that automatically find the best hardware alongside hyperparameter configurations.

This paper makes the following contributions:
\begin{itemize}
  \item We extend Hyperband to handle multiple objectives by relying on non-dominated sorting, which is shown to be more efficient than previously-proposed scalarization approaches.
  \item We propose a new way to achieve transfer learning in the multi-objective setting by building a probabilistic approximation of the Pareto front.
  \item We show in experiments that tuning hardware alongside neural architectures significantly brings down the cost and runtime all else being equal.
  \item We also show that when only optimizing the hyperparameters, 
  our multi-objective method is more efficient than standard multi-objective optimization algorithms.
\end{itemize}


\section{Multi-objective hyperband}
\label{sec:methods}

Hyperband is a multi-fidelity hyperparameter optimization (HPO) algorithm that early stops unpromising configurations during training \citep{Li2018}. 
Random configurations are first evaluated at a small resource level (e.g., a small number of epochs). The fraction of configurations that is performing best is continued and then evaluated at a larger resource level (e.g., a larger number of epochs). This process is repeated until only a few are left and evaluated at the largest resource level. The pseudo-code of Hyperband is provided in the appendix. We present extensions to the sorting and the sampling strategy at the core of Hyperband, allowing us to tackle multi-objective problems.

\begin{figure}
\begin{minipage}{.38\textwidth}
  \centering
\includegraphics[width=0.99\textwidth]{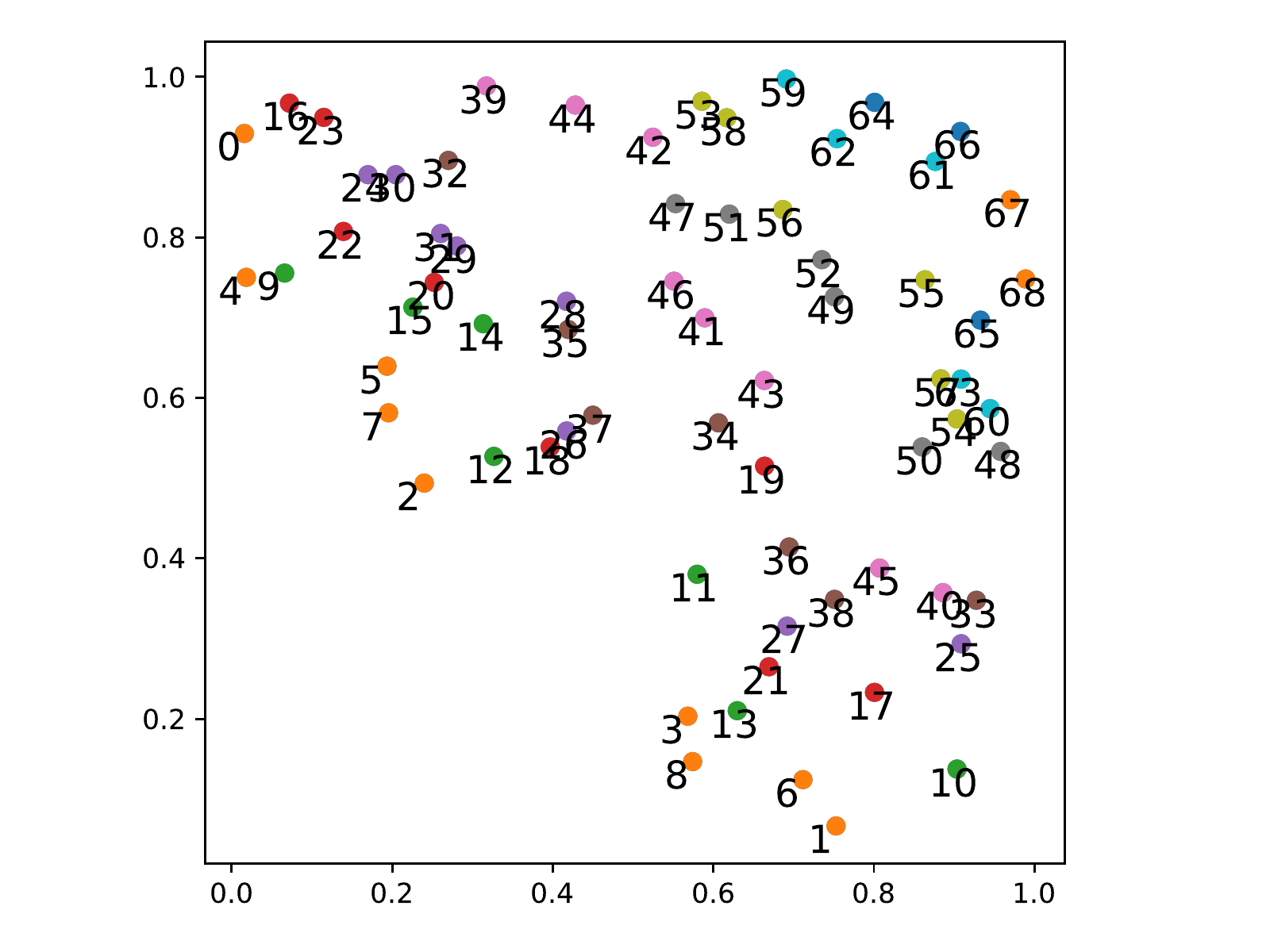}  
\caption{Non-dominated sorting. Labels denotes ranks of different points, colors denotes Pareto front groups. \label{fig:nd-sort}}  
\end{minipage}%
\hspace{1cm}
\begin{minipage}{.6\textwidth}
\center
\includegraphics[width=0.99\textwidth]{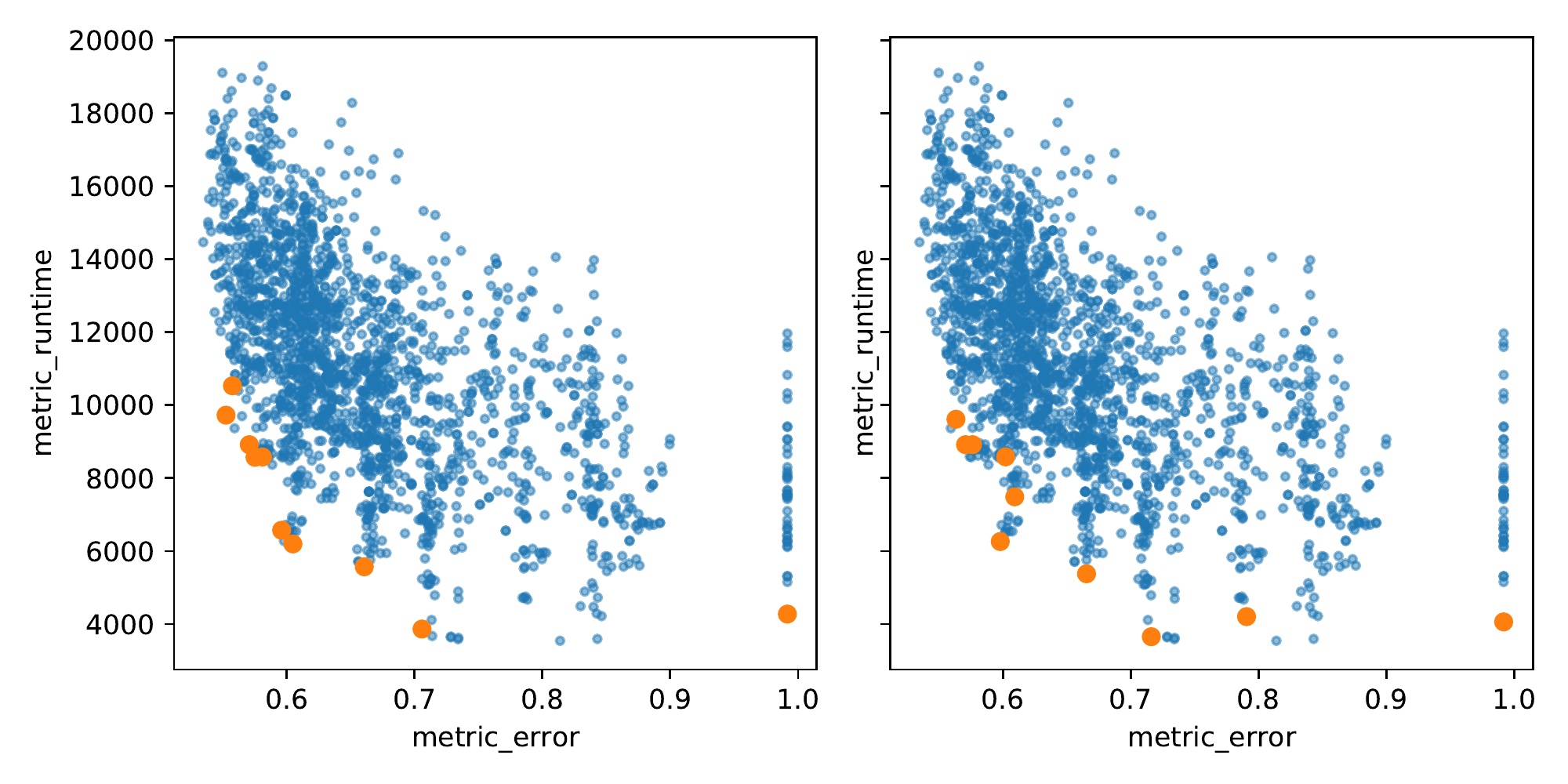}
\caption{Points obtained when sampling surrogates estimates $\tilde{z}(x_1), \dots, \tilde{z}(x_n)$
and picking the first configurations according to non-dominated sorting for two different seeds.
\label{fig:sample}}
\end{minipage}
\end{figure}

\paragraph{Sorting configurations.}
Typically, sorting is done based on accuracy. However, with multiple objectives no strict ordering exists and one has to rely on multi-objective sort. A simple approach is scalarization: a mapping $s: \R^m \to \R$ is used to project the $\nobj$ objectives to a single value as proposed by \cite{momf} and \cite{cruz2020banditbased} for Hyperband. Instead, we propose to use non-dominated sorting~\citep{Emmerich2018}. To the best of our knowledge, we are the first to apply it to multi-objective Hyperband. Figure~\ref{fig:nd-sort} illustrates this approach, which iteratively removes points from the Pareto front and breaks ties via a heuristic priority rule aiming at covering the Pareto front as efficiently as possible. In our case, we use a greedy epsilon-net strategy that first picks the point minimizing the first objective and then iteratively selects the farthest point from the current set. This approach comes with theoretical guarantees on coverage and sparsity~\citep{clarkson2006nearest}. The pseudo-code of the method is given in the appendix.

\ca{This paragraph is providing relatively few technical details; I guess this is the best we can do in 4 pages, but makes the paper not self-contained.}
\ca{More importantly, Figure 2 does not provide any useful information. You either need to drop it or explain what is non-dominant sorting in the caption -- what are the colors and the numbers here?}

\paragraph{Sampling configurations.}
Hyperband samples new configurations at random. 
More efficient strategies build a surrogate model of the optimization objective to direct the sampling \citep{Falkner2018,Tiao2020} or use transfer learning to exploit evaluations from related tasks, such as when tuning the same algorithm on different datasets~\citep{Perrone2019}. We propose a strategy related to the latter suitable for the multi-objective case. We assume a list of $\nhp$ hyperparameters $x_1, \dots, x_\nhp$ are evaluated on $\ntask$ tasks and we denote $y_i^j \in \R^\nobj$ the $\nobj$ objectives recorded for $1\leq i \leq \nhp$ and $1\leq j \leq \ntask$. To sample new configurations, we proceed in three steps. First, we normalize observations with $z_i^j = \psi_j(y_i^j)$, where $\psi_j = \Phi^{-1}~\circ~F_j$ is the normalization used by \cite{salinas20a} with $\Phi$ the Gaussian CDF and $F_j$ the empirical CDF of $y_1^j, \dots, y_{\nhp}^j$. Second, we build a surrogate model of the conditional distribution $p(z | x_i) \approx  \mathcal{N}(\tilde{\mu}_i, \tilde{\sigma}_i^2)$, where 
$\tilde{\mu}_i  = \mathbb{E}_j[z_i^j]$ and $\tilde{\sigma}_i^2 = \mathbb{E}_j[z_i^j - \tilde{\mu}_i]^2$. 
While 
the procedure 
assumes the data is 
collected on a grid, 
it could be adapted through a surrogate model such as an MLP if the hyperparameters are not all evaluated on the different tasks \citep{salinas20a}. Third, we sample the performance for all hyperparameters $\tilde{z}(x_1), \dots, \tilde{z}(x_n)$ according to $p(z | x_i)$ 
and pick the top configurations according to the non-dominated sorting. Figure~\ref{fig:sample} illustrates the sampling of configurations.

\section{Experiments}


\paragraph{Benchmarks.} 
We run experiments on five NAS benchmarks, the first two \nas{} \citep{dong2020nasbench201} and \fcnet{} \citep{Klein2019} contain neural architectures evaluated on multiple datasets for all possible fidelities (number of epochs). For those two datasets, we optimize for hyperparameters minimizing error and runtime. In addition, we use \hwnas{} \citep{li2021hwnasbench}, which contains runtime for all 15625 neural architectures of \nas{} evaluated on 6 different edge devices. We also add cost per hardware which is minimized in addition to error and runtime. Finally, we build two new benchmarks \fcnetcloud{} and \nascloud{} that contain the estimated runtime and cost for all hyperparameters of \nas{} and \fcnet{} on 37 hardware choices available on AWS (containing both cpu and gpu machines; we refer to the appendix for more details). For these three benchmarks, we optimize for error, runtime and cost, with the search space containing the hardware type in addition to hyperparameters. Note that cost and runtime cannot be merged 
as different hardware comes with different cost per hour.

\begin{table}
\center
\scriptsize
\begin{tabular}{l|lll|lll|lll}
\toprule
{} &        \multicolumn{3}{c}{\fcnetcloud} &        \multicolumn{3}{c}{\nascloud} &        \multicolumn{3}{c}{\hwnas} \\
{} &        error &       runtime &           cost &        error &       runtime &          cost &        error &      runtime &          cost \\
\midrule
HB      &  0.26 (100) &  26.74 (100) &  274.13 (100) &  0.56 (100) &  16.65 (100) &  53.39 (100) &  0.55 (100) &  6.71 (100) &   1.68 (100) \\
+Parego &   0.34 (77) &   3.93 (679) &   55.54 (493) &   0.65 (85) &   2.35 (707) &  19.21 (277) &   0.66 (83) &  1.04 (645) &   0.48 (352) \\
+RW     &   0.33 (78) &   4.85 (550) &   54.77 (500) &   0.61 (90) &   3.42 (487) &  21.36 (250) &   0.64 (86) &  1.38 (488) &   0.53 (314) \\
+HV     &   0.31 (84) &   9.21 (290) &   80.54 (340) &   0.58 (95) &  11.16 (149) &  30.61 (174) &   0.61 (91) &  3.25 (206) &   0.86 (193) \\
+ND     &   0.26 (99) &  21.72 (123) &  153.97 (178) &   0.56 (98) &  13.06 (127) &  38.75 (137) &   0.56 (99) &  5.38 (124) &   1.22 (137) \\
+tr     &   0.28 (91) &   4.57 (585) &   4.84 (5666) &  0.54 (102) &   2.38 (700) &   6.62 (806) &  0.55 (100) &  1.11 (605) &  0.09 (1861) \\
+ND+tr  &   0.28 (91) &   4.67 (572) &   4.60 (5964) &  0.56 (100) &   1.99 (836) &   6.52 (819) &   0.56 (99) &  0.91 (736) &  0.08 (2115) \\
\bottomrule
\end{tabular}
\caption{Error, runtime in hours and cost in dollars when training Hyperband and multi-objective variants for hardware-aware benchmark. Percentage improvement over \HB{} is shown in parenthesis. \label{tab:benchmark-hw-aware}}
\end{table}

\paragraph{Results.}
Table \ref{tab:benchmark-hw-aware} and \ref{tab:benchmark} report the average objectives obtained after running Hyperband (\HB{}) and the following multi-objective variants for 30 seeds. We compare with different scalarizations, namely Parego (\HBparego{}) \citep{parego}, the random-weight approach (\HBrw{}) used for multi-objective multi-fidelity optimization by \cite{momf}, the scalarization of \cite{Golovin2020} which enjoys theoretical regret guarantees, and the non-dominated sort approach we propose (\HBND{}). In addition, we report results of HB when using the transfer learning-based sampling described in the previous section (\HBtr{}) and when using in addition non-dominated sorting (\HBNDtr{}) in contrast to ranking only with accuracy. 

\begin{wraptable}{r}{7cm}
\center
\scriptsize
\begin{tabular}{l|ll|ll}
\toprule
{} &        \multicolumn{2}{c}{\nas}    &        \multicolumn{2}{c}{\fcnet} \\
{} &        error &       runtime &        error &      runtime \\
\midrule
HB      &  0.26 (100) &  14.86 (100) &  0.56 (100) &  5.39 (100) \\
+Parego &   0.34 (77) &   3.15 (471) &   0.69 (80) &  2.78 (193) \\
+RW     &   0.33 (77) &   3.59 (413) &   0.62 (90) &  4.06 (132) \\
+HV     &   0.30 (86) &   5.79 (244) &   0.57 (97) &  4.77 (112) \\
+ND     &   0.26 (99) &  12.30 (120) &   0.56 (99) &  4.91 (109) \\
+tr     &  0.24 (109) &   4.47 (332) &  0.56 (100) &  4.97 (108) \\
+ND+tr  &  0.24 (108) &   3.91 (379) &  0.56 (100) &  4.76 (113) \\
\bottomrule
\end{tabular}
\caption{Methods comparison when tuning hyperparameters minimizing error and runtime.
Percentage improvement over \HB{} is shown in parenthesis.
 \label{tab:benchmark}}
\end{wraptable} 

\begin{figure}[t]
\center
\includegraphics[width=0.4\textwidth]{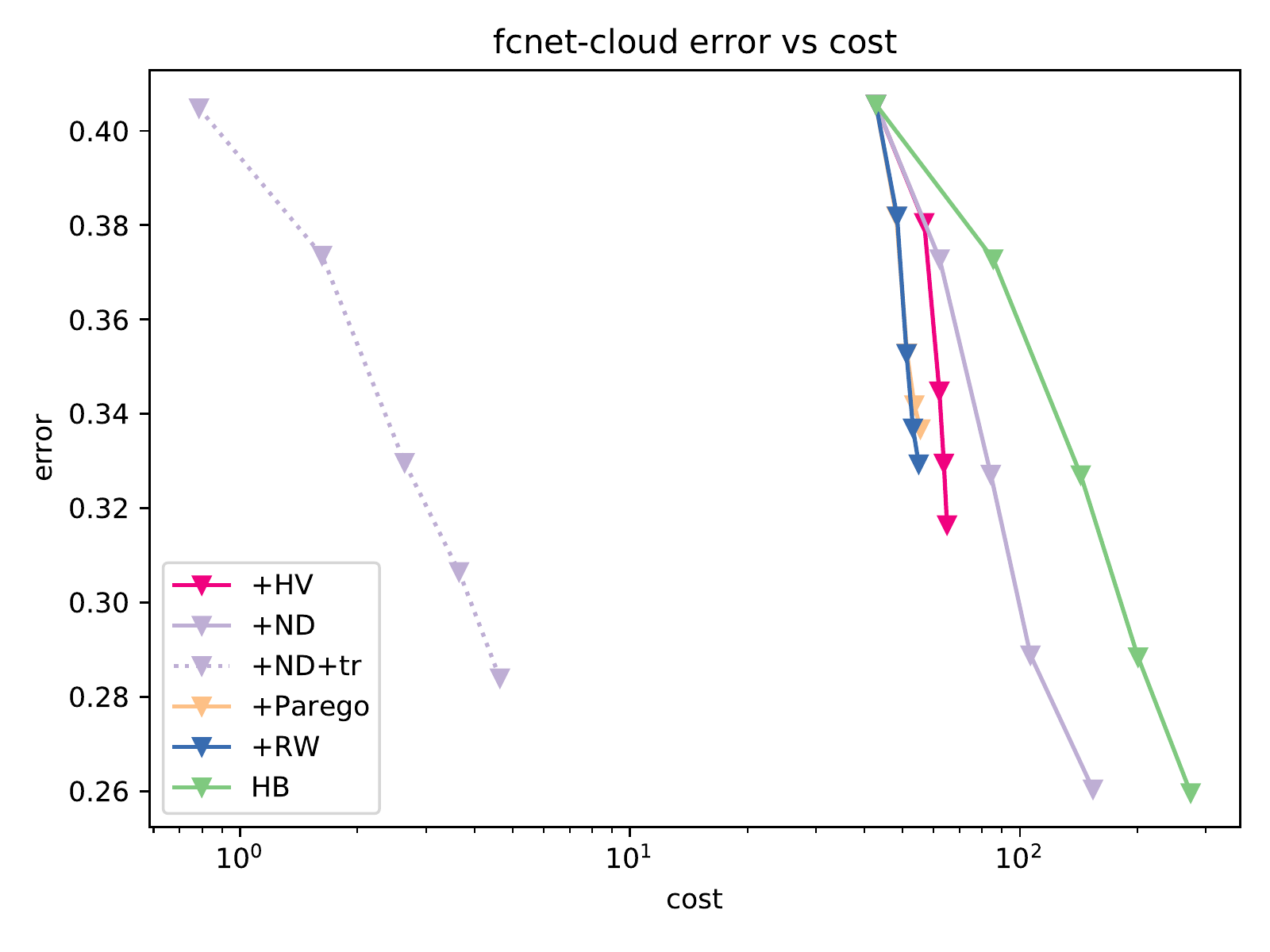}
\includegraphics[width=0.4\textwidth]{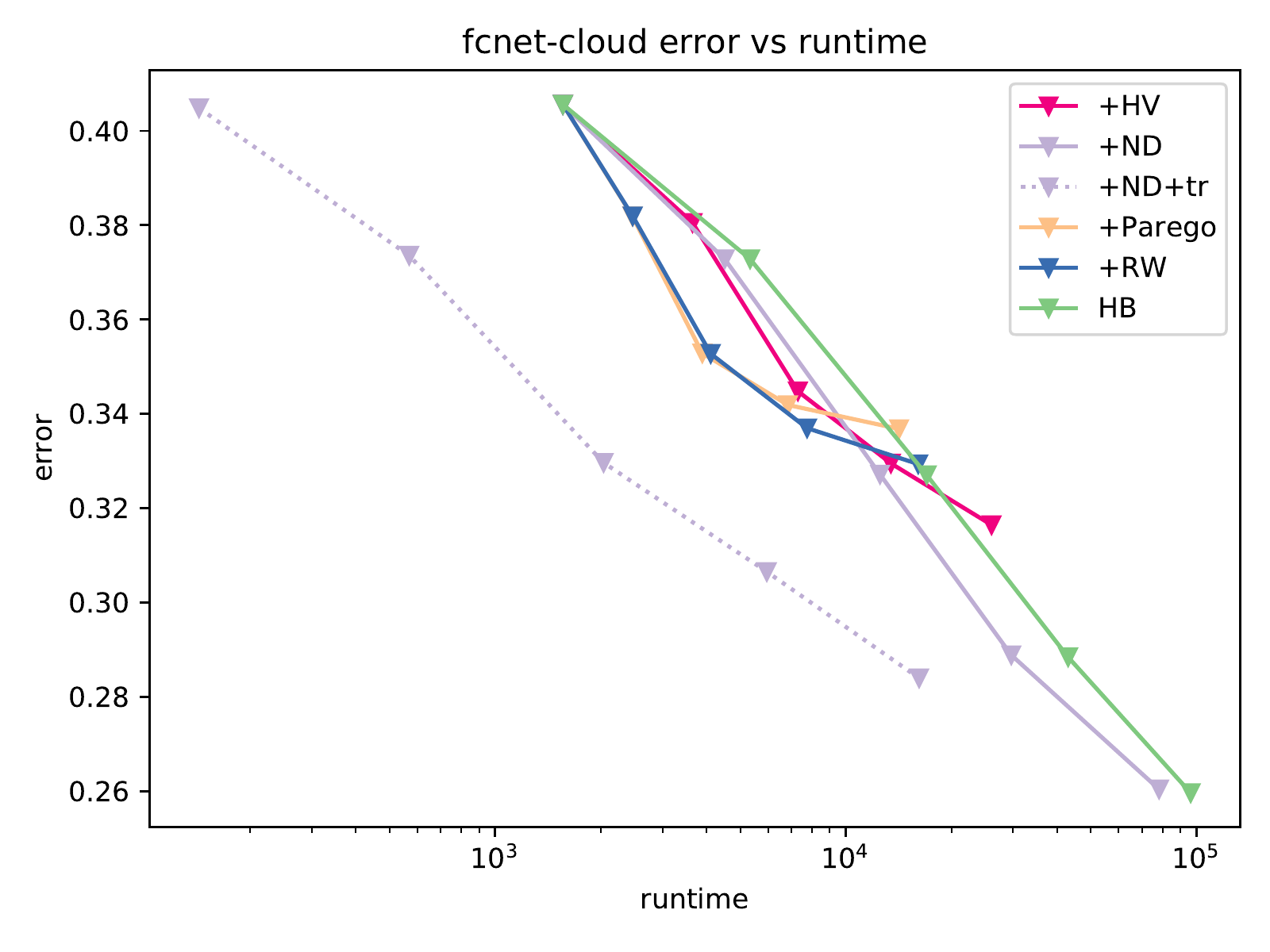}
\includegraphics[width=0.4\textwidth]{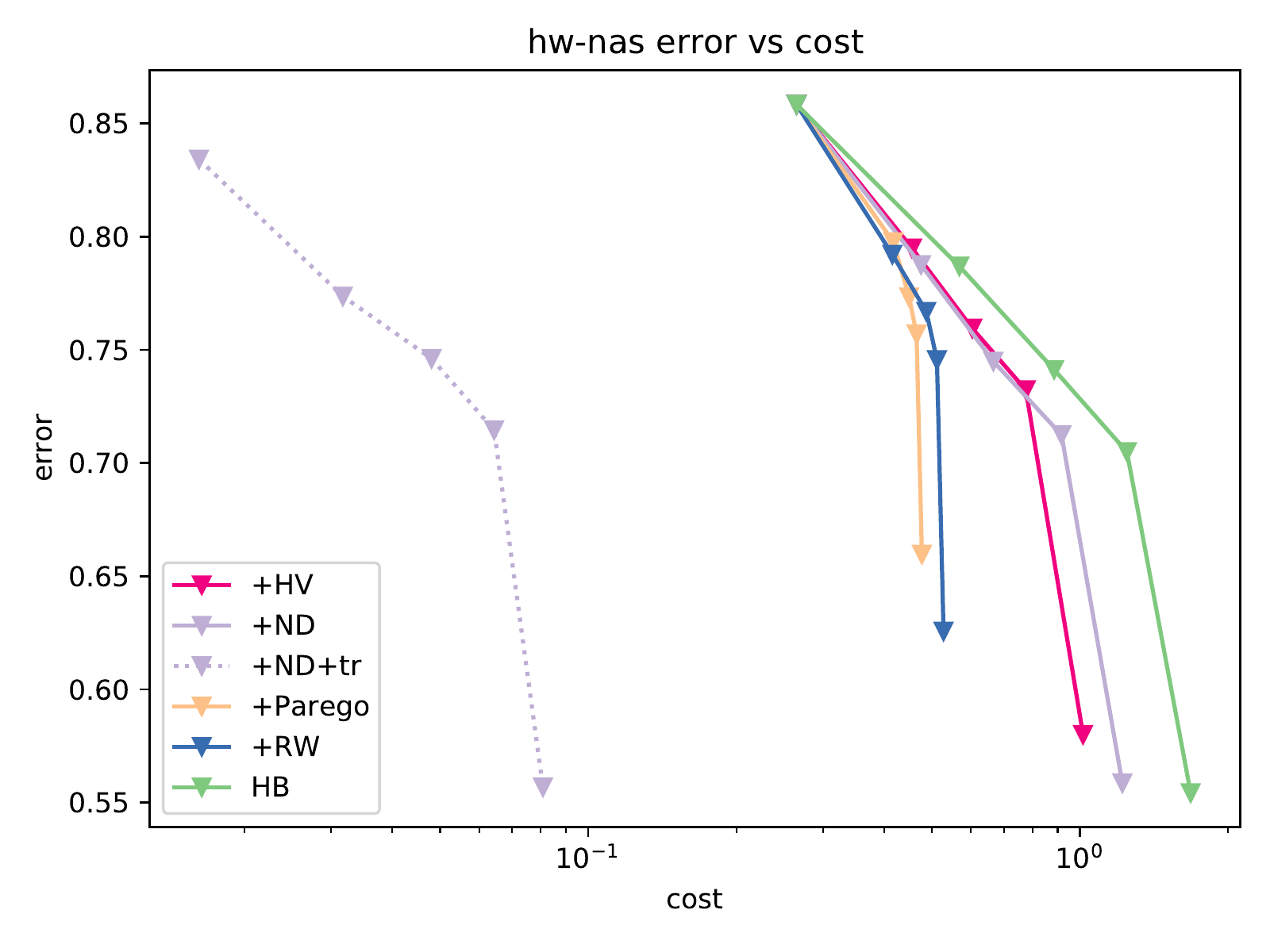}
\includegraphics[width=0.4\textwidth]{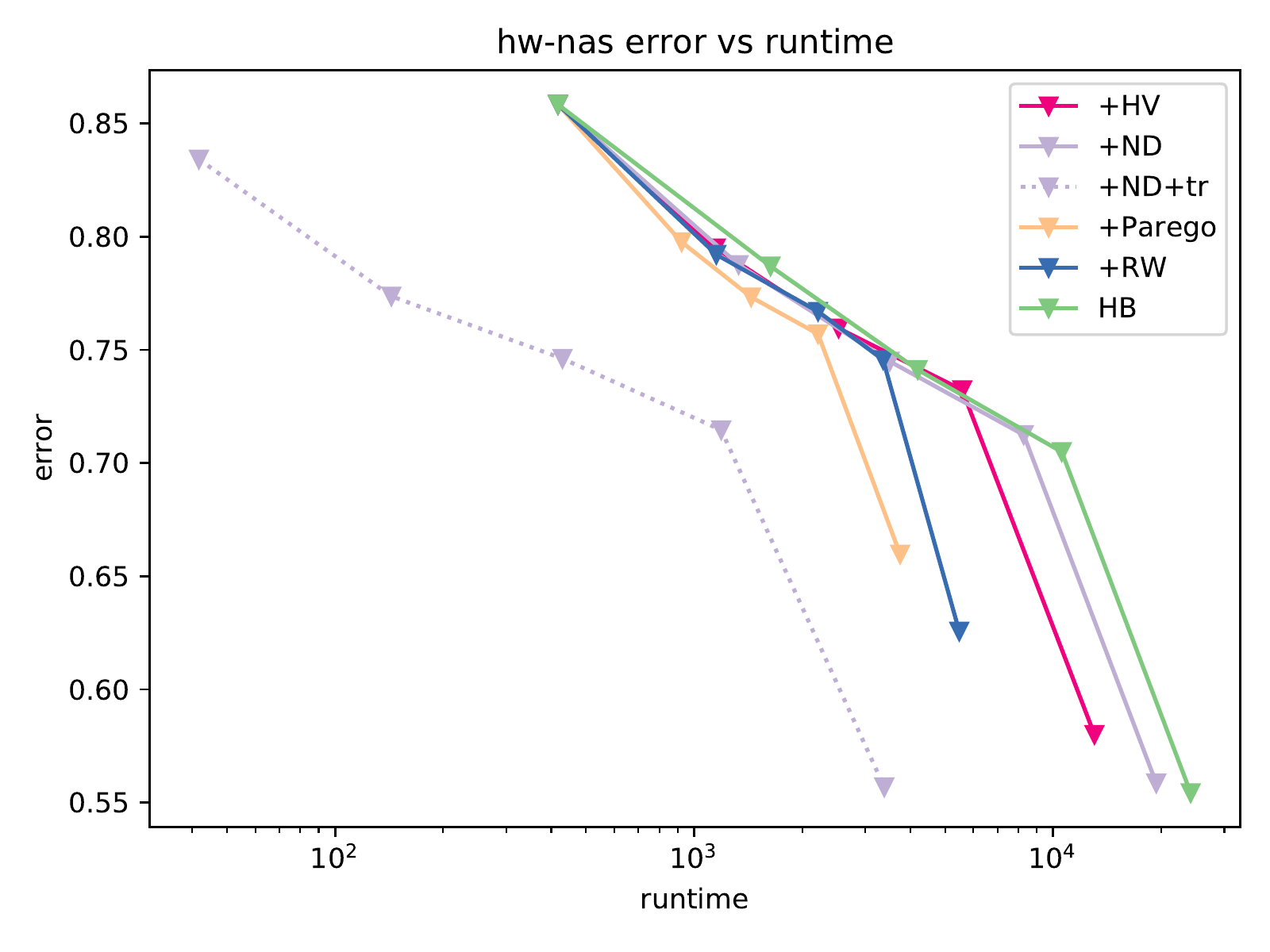}
\caption{Error and cost (left, in dollars) or runtime (right, in seconds) when running HB with different strategies for \fcnetcloud{} (top) and \hwnas{} (bottom). Non-dominated sorting and transfer learning allow us to get similar final accuracy while significantly reducing cost and runtime, especially when we leverage observations from related tasks. \label{fig:hb-conv}}
\end{figure}

Table \ref{tab:benchmark} shows that multi-objective non-dominated sorting reaches very similar errors as when minimizing only the accuracy, while also reducing the runtime. In the case where we only the hyperparameters, and thus not the hardware, we obtain runtime speed-ups of 20\% and 10\%, respectively, for \fcnet{} and \nas{} while maintaining the same accuracy. In the benchmarks where we also tune the hardware (Table \ref{tab:benchmark-hw-aware}), the gains are all greater than 20\% while we maintain a very similar accuracy. Other scalarization approaches methods save runtime/cost but a large decrease of accuracy is observed. We conjecture this is due to their independent ranking of configurations, which is outperformed by dominated sorting. Finally, the multi-objective transfer learning approach that we propose (\HBNDtr{}) enables to save several order of magnitude in runtime and cost as can be seen in Figure~\ref{fig:hb-conv} which compares all baselines in terms of average error against runtime/cost at all resource levels on \nascloud{} and \hwnas{}.

 \section{Conclusion} 
 \label{sec:conclusion}
We investigated the impact of hardware selection in the context of multi-fidelity hyperparameter optimization, showing that automatically selecting it brings large cost and runtime savings. 
The dramatic impact could be further pronounced when also considering distributed training or float precision. In addition, our method 
yields some improvements when searching only the hyperparameters. 
Future work could include user constraints, such as not exceeding a cost budget, as well as model feasibility constraints, such as avoiding configurations prone to out of memory errors.


\bibliography{iclr2021_conference}
\bibliographystyle{iclr2021_conference}

\newpage

\appendix
\section{Appendix}

For \nas{}, we use \emph{ImageNet16-120} as test-task and use \emph{Cifar10} and \emph{Cifar100} to build a 
transfer-learning surrogate model. For \fcnet{}, we use \emph{protein\_structure} as test-tasks and use the 3 remaining datasets to build the transfer-learning model. Those two datasets were chosen as they are the most difficult as the tasks are the most distinct from other available tasks, we will consider evaluating every combination in future work. 

\subsection{Cost estimation of \hwnas{}, \nascloud{} and \fcnetcloud{}}
\paragraph{\nascloud{} and \fcnetcloud{}.} As evaluations are only made given a single hardware choice for \nas{} and \fcnet{}, we estimate the runtime that would have been obtained on different hardware. To this end, we measure the runtime per batch $r_h$ of a similar model (resnet for \nas{}, MLP for \fcnet{}) on 37 different instance types $h\in \HW$ on AWS. The runtime of a configuration $x\in \HP$ for a hardware $h\in \HW$ is then estimated as $r(x, h) = r^{\text{original}}(x) \frac{r_h}{r_{h_{\text{ref}}}},$
where $r^{\text{original}}(x)$ is the original runtime measurement of the benchmark for the hyperparameter $x$ and $h_{\text{ref}}$ is a instance chosen to be close to the one used in the benchmark (\texttt{p2.xlarge} for \nas{} and \texttt{ml.c4.4xlarge} for \fcnet{}). 

In addition, we add a dollar cost objective $c(x, h) = \gamma_h r(x, h)$ based on the estimated runtime, where $\gamma_h$ is the cost per second of the hardware $h$. In short, we are optimizing three objectives: error $e(x, h)$, runtime $r(x,h)$ and cost $c(x,h)$. The search space is $(x, h) \in \HP =  \R^d \times \HW$, where $\HW$ denotes the discrete set of possible instances and where the set of original hyperparameter is assumed in $\R^d$. 

\paragraph{\hwnas{}.}

We estimate the total cost for each hardware of \citep{li2021hwnasbench} as follow:

\begin{table}[h]
\center
\begin{tabular}{c|c|c|c|c|c|c}
\toprule
gpu1080 & edgegpu & raspi4& edgetpu & pixel3 & eyeriss &fpga \\
\midrule
800\$ &  499\$ & 99.99\$ &  129.99\$ & 140\$ & 2500\$ & 2500\$ \\
\bottomrule
\end{tabular}
\end{table}

The runtime of $r(x, h)$ of a hyperparameter $x$ and a hardware $h$ is estimated with 
$$r(x, h) = r_{\nas{}}(x) \frac{\text{lat}_{\hwnas{}}(x, h)}{\text{lat}_{\nas{}}(x)}$$ where $r_{\nas{}}(x)$ denotes the runtime measured in \nas{} for the hyperparameter $x$, $\text{lat}_{\hwnas{}}(x, h)$ the latency measured in \citep{li2021hwnasbench}  and $\text{lat}_{\nas{}}(x)$ denotes the latency measured in \nas{}.

Total costs of different hardware options were estimated by communication with the author of \citep{li2021hwnasbench} except for eyeriss whose price is unknown and estimated by us as the hardware is not on the market. To get the cost of a job $c(x, h)$ of an hyperparameter $x$ and a hardware $h$, we multiply the runtime $r(x, h)$ by the total hardware cost divided by $3600 \times 24 \times 200$ (which assumes the hardware would have to be changed after $200$ days).

\subsection{Pseudo-code}

\paragraph{Hyperband.}
At its core, the multi-fidelity algorithm we propose builds on successive halving (SH), which we briefly describe.  Let $f(x, r)$ be the target objective as a function of its hyperparameters $x$ and the fidelity level $r$ (e.g., the number of epochs at which $f(x)$ is evaluated). The overall procedure is described in Algorithm~\ref{alg:hb}. First, all configurations are evaluated at the smallest resource level. Then, only the best $1/\eta$ configurations are kept in the set of configurations $C$ (line 5). The fidelity parameter is then increased by a factor of $\eta$. This
is repeated until the maximum budget is reached. SH is simple to implement and can be an efficient baseline, especially compared to standard HPO.  However, selecting the number of number of configurations to evaluate is not easy in practice. For this reason, SH has been extended to Hyperband, which adds an outer iteration over different values of $n$ and different budget, we give the pseudo-code from the method from \citep{Li2018} on Alg. \ref{alg:hb}.

Importantly, the method only depends on sampling (line 5) and sorting configurations (line 10). The sorting can be extended with any scalarization method or with the non-dominated sort whose pseudo-code is given in Alg.~\ref{alg:nd}.

\begin{algorithm}[t] 
\small
\textbf{Input: } $R, \eta$ (default to 3) 

$s_\text{max} = \log_\eta(R)$, $B = (s_\text{max} + 1) R$ \;

\For{$s \in \{s_\text{max}, \dots, 0\}$}{

$n = [\frac{B}{R} \frac{\eta^s}{s+1}], r = R \eta^{-1}$ \;
$ X = \text{sample\_configuration}(n)$ \;

\For{$i \in \{0, \dots, s\}$}{
$n_i = [n \eta^{-i}]$ \;
$r_i = r \eta^i$ \;
$L = \{f(x, r_i)~|~\forall X\in X\}$ \;
$X = \text{top}_k(X, L, [n_i/\eta])$ \;
}
}

 \caption{Pseudo-code for Hyperband.}
\label{alg:hb}
\end{algorithm}

\paragraph{Baseline details.}

We use the following scalarizations $s: \R^\nobj \to \R$ as baseline to pick the $\text{top}_k$ configurations:
\begin{itemize}
  \item Linear (\HBrw{}) \citep{momf}: $s(y) = \lambda^T y$ where $\lambda$ is drawn randomly on a simplex when the function is called (we tried also searching for different values $\lambda$ but the method perform poorly).
  \item Parego (\HBparego{}) \citep{parego}: $s(y) = \max_i \lambda_i y_i + \rho \sum_i \lambda_i y_i$ where $\rho > 0$ is a hyperparameter set to a small value (0.05 in  \citet{parego} and our implementation) and $\lambda$ is drawn randomly on a simplex when the function is called.
  \item Hypervolume (\HBhv{}) \citep{Golovin2020}
$s(y) = \min_i (\max(0, y_i/\lambda_i))^\nobj$ 
where $\lambda$ is drawn randomly from the unit positive sphere when the function is called.
\end{itemize}

All objectives are centered with mean $0$ and variance $1$ with available observations. As \citet{Golovin2020} requires positive data, we also substract the minimum objective to map to positive values only for this scalarization.

The method we propose is non-parametric and both the non-dominated sort and the transfer learning approach have no hyperparameters (as the probabilistic surrogate of the Pareto front is built with empirical mean and variance). We use values in $[0, 29]$ for the seeds. The runtime of every method is bellow 1s for all benchmarks considered including running Hyperband for all iterations (excluding time to load tabular data).

\paragraph{Non-dominated sorting and epsilon-net pseudo-code.}
Alg.~\ref{alg:nd} and Alg.~\ref{alg:sort} give pseudo-code for non-dominated and epsilon-net sort, respectively.

\begin{algorithm}
\small
\textbf{Input: } Set of objectives $y \in \R^{\nhp \times \nobj}$ \;
\textbf{Output: } Points sorted $\in \R^{\nhp \times \nobj}$ \;

\uIf{$\nhp \leq 1$}{
  return y \;
}
\Else{
\tcc{compute points of $y$ that are in the Pareto front}
$ P = \text{Pareto}(y)$ \;

\tcc{use heuristic to break tie among points of $P$, call recursively on remaining points}
\Return $\text{Sort}(P) + \text{ND-sort}(y \setminus P)$ \;
}

 \caption{Pseudo-code for non-dominated sort.}
\label{alg:nd}
\end{algorithm}

\begin{algorithm}
\small
\textbf{Input: } Set of objectives $y \in \R^{\nhp \times \nobj}$ \;
\textbf{Output: } Points sorted $\in \R^{\nhp \times \nobj}$ \;

\uIf{$\nhp \leq 0$}{
  return y \;
}
\Else{

\tcc{initialize with the point of $y$ which has the lowest first coordinate}
$i_1 = \text{argmin}y_{., 1}$ \;
\For{$k \in [2, \nhp]$}{

\tcc{pick point furthest away from the previously selected points}
$i_k = \text{argmax}_{i\not\in \{i_1, \dots, i_{k-1}\}} d(y_i, \{y_{i_1}, \dots, y_{i_{k-1}}\})$ \;
}
\Return $y_{i_1}, \dots, y_{i_\nhp}$ \;
}
\caption{Pseudo-code for Sorting in the non-dominated sort (epsilon-net). \label{alg:sort}} 
\end{algorithm}



\end{document}